\documentclass[letterpaper, 10 pt, conference]{ieeeconf}  

\IEEEoverridecommandlockouts                              

\usepackage[utf8]{inputenc}
\usepackage{amsmath}
\usepackage{amssymb}             
\usepackage{amsfonts}            
\usepackage{mathrsfs}            
\usepackage{graphicx}
\usepackage{float}
\usepackage{booktabs}
\usepackage{times}
\usepackage{multirow} 
\usepackage{wrapfig}
\usepackage{cuted}
\usepackage{capt-of} 
\usepackage[caption=false,font=footnotesize]{subfig}
\usepackage{booktabs, tabularx, makecell}

\usepackage{stfloats}
\usepackage{makecell}

\title{\LARGE \bf
Learning Generalizable Robot Policy with Human \\ Demonstration Video as a Prompt
}

\begin{document}


\author{
Xiang Zhu$^{1,2*}$, 
Yichen Liu$^{1*}$, 
Hezhong Li$^{2}$, 
Jianyu Chen$^{1,2\dagger}$%
\thanks{$^{1}$Tsinghua University, China. 
$^{2}$Shanghai Qi Zhi Institute, China. 
$^{*}$Equal contribution. 
$^{\dagger}$Corresponding author.}
}
\maketitle

\bstctlcite{IEEEexample:BSTcontrol}


\begin{figure*}[h]
    \centering
    \hspace*{-7mm}
    \includegraphics[width=1.00\linewidth]{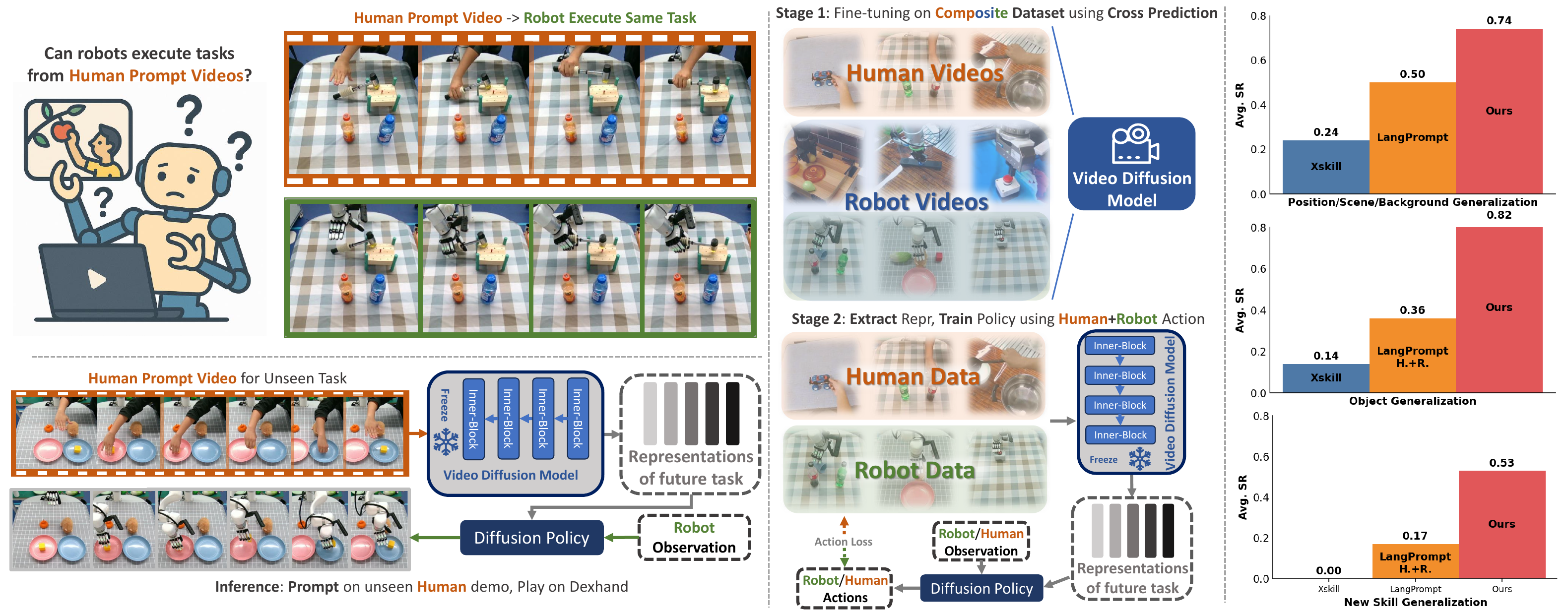}
    \vspace*{-3mm}
    \captionof{figure}{First, fine-tune a video diffusion model on diverse datasets to obtain informative representations. In the second stage, use a video generation model to extract information from human prompt videos for skill learning with both human and robot data. Finally, during inference, the model prompts on unseen human demos to perform tasks based on the human input.}
    \label{fig:main_exp}
    \vspace*{-6mm}
\end{figure*}


\begin{abstract}

Recent robot learning methods commonly rely on imitation learning from massive robotic dataset collected with teleoperation. When facing a new task, such methods generally require collecting a set of new teleoperation data and finetuning the policy. Furthermore, the teleoperation data collection pipeline is also tedious and expensive. Instead, human is able to efficiently learn new tasks by just watching others do. In this paper, we introduce a novel two-stage framework that utilizes human demonstrations to learn a generalizable robot policy. Such policy can directly take human demonstration video as a prompt and perform new tasks without any new teleoperation data and model finetuning at all. In the first stage, we train video generation model that captures a joint representation for both the human and robot demonstration video data using cross-prediction. In the second stage, we fuse the learned representation with a shared action space between human and robot using a novel prototypical contrastive loss. Empirical evaluations on real‑world dexterous manipulation tasks show the effectiveness and generalization capabilities of our proposed method.

\end{abstract}

\section{Introduction}
Traditional robot learning methods typically train language-conditioned policies on large datasets collected through teleoperation\cite{Brohan2022RT1RT, Brohan2023RT2VM}. While effective for known tasks, this paradigm faces two fundamental limitations when encountering novel tasks. First, language instructions, though intuitive for humans, provide only categorical information and lack the rich spatial and temporal details crucial for physical manipulation. Second, adapting to new tasks usually requires collecting additional robot demonstrations, a process that is both time-consuming and expensive due to the complexity of teleoperation systems.

Humans, in contrast, can efficiently acquire new skills simply by observing others perform tasks. This observation suggests that visual demonstrations may offer a more natural and information-rich medium for teaching robots. Videos inherently capture not just what task to perform, but also how to perform it, including critical aspects like object relationships, motion trajectories, and timing. Moreover, human demonstration videos are significantly more scalable to acquire compared to robot data, whether captured in laboratory settings or obtained from existing online resources.


Recent research has only begun to explore this direction, and existing approaches exhibit notable limitations. For instance, EgoMimic \cite{kareer2024egomimic} primarily focuses on single-task scenarios. UniSkill \cite{kim2025uniskill} also incorporates human videos as prompts, but these are only utilized in the Stage 1 representation learning phase to learn action embeddings, while the Stage 2 policy learning stage does not leverage them; its goal is to highlight the effect of human videos on simple generalization through representation learning. Similarly, Gen2Act \cite{bharadhwaj2024gen2act} and Human2Robot \cite{Xie2025Human2RobotLR} both integrate human videos into training, but the number of human demonstrations involved is relatively limited, leaving them insufficient for achieving robust generalization when combined with sparse robot demonstrations. Overall, the field still lacks a general framework that effectively leverages human videos together with sparse robot demonstrations to enable challenging forms of generalization.

We propose a two-stage human-prompted learning framework that combines robotic datasets with human demonstration data to address challenges in task learning.
In the first stage, we use a video generation model that receives a prompt video of a human performing a task and an image of the robotic hand. The model generates a video of the robot performing the task, embedding embodiment transfer information through a cross-prediction strategy. This helps the model learn a representation that captures the task, context, and target modality effectively.
In the second stage, we fine-tune the representation using a diffusion policy, leveraging abundant human demonstration data together with robot data. A unified action space bridges the gap between the two modalities, while a cluster-based loss enhances skill separation and multi-skill imitation performance.
Experiments on real-world tasks demonstrate the framework’s effectiveness in improving human-robot interaction and manipulation. 


Our core contribution is achieving elementary generalization ability to new tasks and novel skills from human prompt video by proposing a novel learning framework that effectively leverages extensive human demonstration videos. Concretely, our contributions are threefold: (1) a cross-embodiment learning strategy based on cross-prediction for representation learning, (2) joint training with human video–action pairs to enhance policy generalization, and (3) the ProtoDiffusion Contrastive Policy objective to improve representation quality. Throughout the paper, we focus on evaluating generalization across three progressive levels: basic variation, novel objects, and novel skills, under the setting of human video prompting.

\section{Related Works}
\subsection{Learn from Human Videos}

Recent advances in robot manipulation have increasingly utilized human video data to enhance both dexterous and gripper-based manipulation. In dexterous manipulation, works like \cite{chen2024vividex, qin2022dexmv, wang2024dexcap} focus on fine-grained control of multi-fingered systems, while \cite{kannan2023deft} integrates affordance cues. For gripper manipulation, end-to-end video-conditioned policies such as \cite{jain2024vid2robot, yuan2024general, ju2024robo, xu2023xskill} translate visual cues into actionable policies.
Approaches using paired human–robot demonstration data, including \cite{wang2023mimicplay}, \cite{smith2019avid, xiong2021learning}, \cite{wen2023any}, \cite{gu2023rt}, \cite{kareer2024egomimic}, \cite{Xie2025Human2RobotLR}, and \cite{ye2024latent}, address the domain gap by linking human actions with robot trajectories. More recently, generative video techniques like \cite{liang2024dreamitate}, \cite{du2023learning}, \cite{bharadhwaj2024gen2act}, and \cite{shridhar2024generative} leverage video synthesis and textual cues to generate visuomotor policies. These studies highlight the growing trend of using human video demonstrations, paired data, and generative methods to create more adaptable and robust robot manipulation policies.


\vspace*{-2ex}
\subsection{Video as Prompt for Robotic Learning}

Recent studies \cite{chang2023one, xiong2023robotube} have increasingly leveraged human demonstration videos to guide robot learning. For instance, \cite{xiong2021learning} mitigates the embodiment mismatch between humans and robots by converting human videos into robot-centric demonstrations via unsupervised domain adaptation and keypoint extraction. \cite{chane2023learning} enables zero-shot generalization by conditioning robot policies on pretrained video embeddings. Similarly, \cite{jain2024vid2robot} employs cross-attention transformers to map human videos to robot actions, while \cite{qian2024contrast} improves sample efficiency and generalization through contrastive learning, imitation, and limited adaptation. \cite{xu2023xskill} focuses on cross-embodiment skill discovery to obtain transferable representations. Building on precisely aligned human–robot video pairs in the H\&R dataset, \cite{Xie2025Human2RobotLR} introduces a two-stage framework centered on video generation, achieving fine-grained temporal alignment and substantially enhancing both generalization and one-shot capabilities. 
In contrast, \cite{kim2025uniskill} does not rely on paired data but instead proposes a cross-embodiment representation learning framework that jointly leverages large-scale human videos and robot demonstrations, demonstrating stronger generalization in cross-embodiment tasks.

\subsection{Diffusion Models for Robot Policy Learning}

Diffusion models have shown great success in generative computer vision \cite{Ho2020DenoisingDP, Song2020ScoreBasedGM}, leading to their adaptation in robotic policy learning. Pioneering works \cite{Chi2023DiffusionPV, Pearce2023ImitatingHB, Reuss2023GoalConditionedIL} demonstrated their ability to generate denoised robot actions and capture multi-modal behavior distributions. Scaling efforts, such as \cite{Team2024OctoAO}, showcased generalization across diverse robotic platforms with transformer-based diffusion policies pretrained on the Open X-Embodiment dataset. Models like MDT \cite{Reuss2024MultimodalDT} and RDT-1B \cite{Liu2024RDT1BAD} use transformer-based diffusion models, replacing the traditional U-Net. RDT-1B further unifies action representations across robots and incorporates multi-robot data for bimanual manipulation.


Diffusion-based policies can model multimodal action distributions in high-dimensional spaces \cite{Chi2023DiffusionPV, wang2024diffusion, jain2024sampling}. \cite{li2024learning} augments these models with unsupervised clustering and intrinsic rewards to maintain multiple behavior modes, while \cite{zhang2024entropy} adds an entropy regularizer to enhance robustness. Building on these approaches, we propose injecting SPCL’s \cite{li2020prototypical, mo2022siamese} prototypical contrastive loss into diffusion policy training, leveraging task category labels as supervised prototypes. Thereby sharpening policy representations along known task dimensions and enhancing multi‐task imitation performance. Our method can be framed as "in-context learning", where human demonstrations serve as the contextual basis for the policy to achieve zero-shot transfer of unseen robotic skills.


\section{METHODS}
In this section, we detail our framework by outlining its two-stage training process. Our goal is to enable the agent to learn and extract meaningful features and representations from human demonstration videos to perform specific tasks. The algorithm is divided into two stages. In the first stage, we develop a robust representation from a large collection of human data, robotic gripper data, and a small amount of dexterous hand data. This representation captures the target modality, the intended task, and the scene context. In the second stage, using this solid representation, we learn human manipulation patterns from abundant human data and a limited set of high-quality dexterous hand operation data, transferring these patterns to the dexterous hand to achieve better generalization with only a small amount of valuable dexterous hand data.

\begin{figure*}
    \centering
    \vspace*{-2mm} 
    \includegraphics[width=1.00\linewidth]{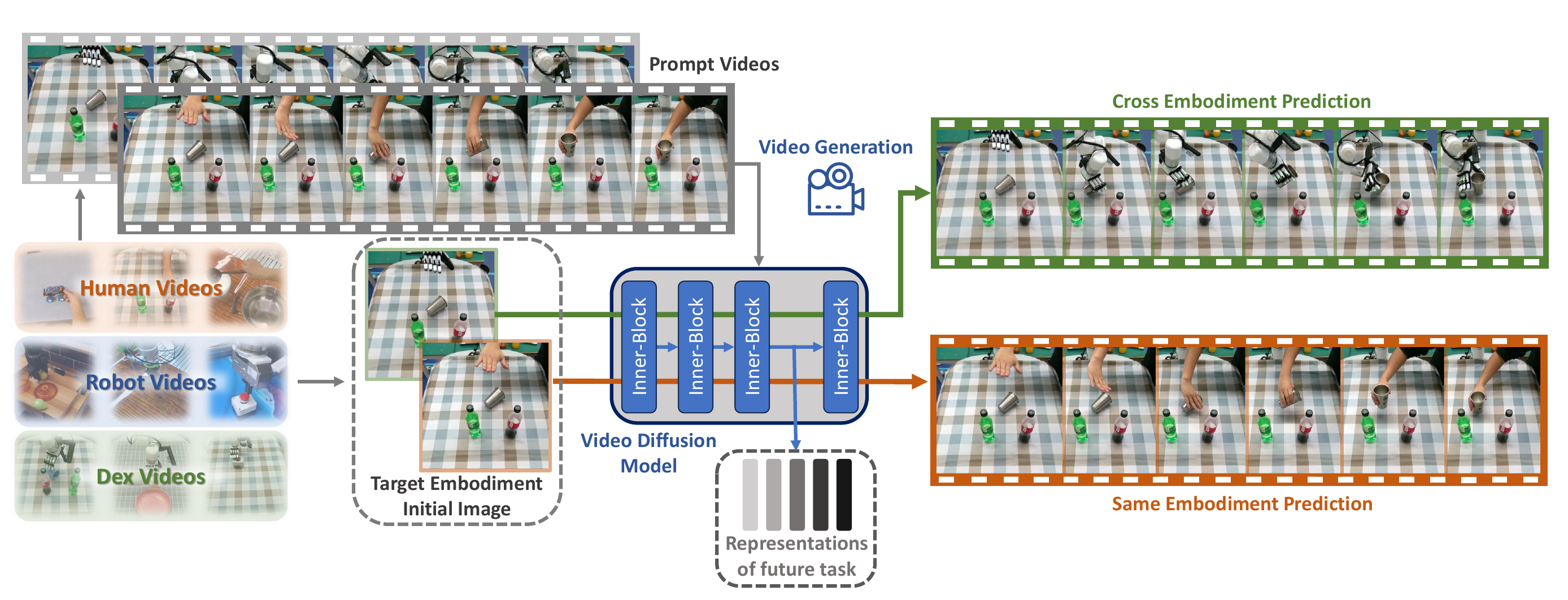}
    \vspace*{-5mm}
    \captionof{figure}{\textbf{Stage 1}: Our video generation model takes as input a task demonstration from a source embodiment together with the initial scene of a target embodiment, and outputs a video in which the target embodiment performs the same task, thereby enabling embodiment transfer.}
    \label{fig:stage1}
    \vspace*{-5mm}
\end{figure*}

\subsection{Augment \textbf{V}ideo \textbf{G}eneration by \textbf{C}ross \textbf{P}rediction(VGCP)}
\label{sec:CrossPrediction}

Recent advancements in video generation models have leveraged extensive online video datasets containing prior knowledge about physical world dynamics. However, these models lack data relevant to robotic manipulation. To address this, we fine-tune an existing model with a custom dataset focused on robotic and object manipulation. Our goal is to enable agents to perform tasks based on human prompt videos, particularly complex dexterous hand operations. Therefore, the video generation model is trained to include details on human manipulation, object motion, scene context, and affordances. Our dataset consists mainly of human manipulation videos, supplemented by robotic gripper and self-collected dexterous hand videos. While gripper videos differ from dexterous hand operations, they provide valuable information on object movement and scene understanding, enhancing the model’s performance.


\textbf{Cross-Prediction:} To further leverage our data and enhance the quality of the learned representation, we propose a method called cross-prediction. As shown in Fig.\ref{fig:stage1}, our video generation model receives a video prompt showing a task performed by a \textbf{source embodiment} and an initial scene for a \textbf{target embodiment}, then generates a video of the target embodiment performing the task, effectively transferring the embodiment. For example, given a prompt video of a human grasping a cup, the model produces a video of a dexterous hand performing the same action. During training, we choose cross-prediction with probability $P$ (using different \textbf{source} and \textbf{target} embodiments) or normal-prediction with probability $1-P$ (using the same embodiment). This approach embeds embodiment transfer information into the video generation model, further improving the transferability of the learned representation. Our cross-prediction method randomly selects different source (s) and target (t) embodiments with probability $P$, and identical ones with probability $1-P$. When the two embodiments are the same, the process mirrors typical video generation models that generate subsequent frames from an initial frame. Our goal is for the model to learn the modality transfer between human and robot, while preserving existing knowledge of their manipulation. We believe that this approach enables the model to capture a video prompt representation that encompasses the skills used by the source embodiment, the manipulated objects, and some environmental context. The fine-tuned model will then be frozen during the second stage of training.

\begin{figure*}[t]
    \centering
    \vspace*{-1mm}
    \includegraphics[width=0.65\linewidth]{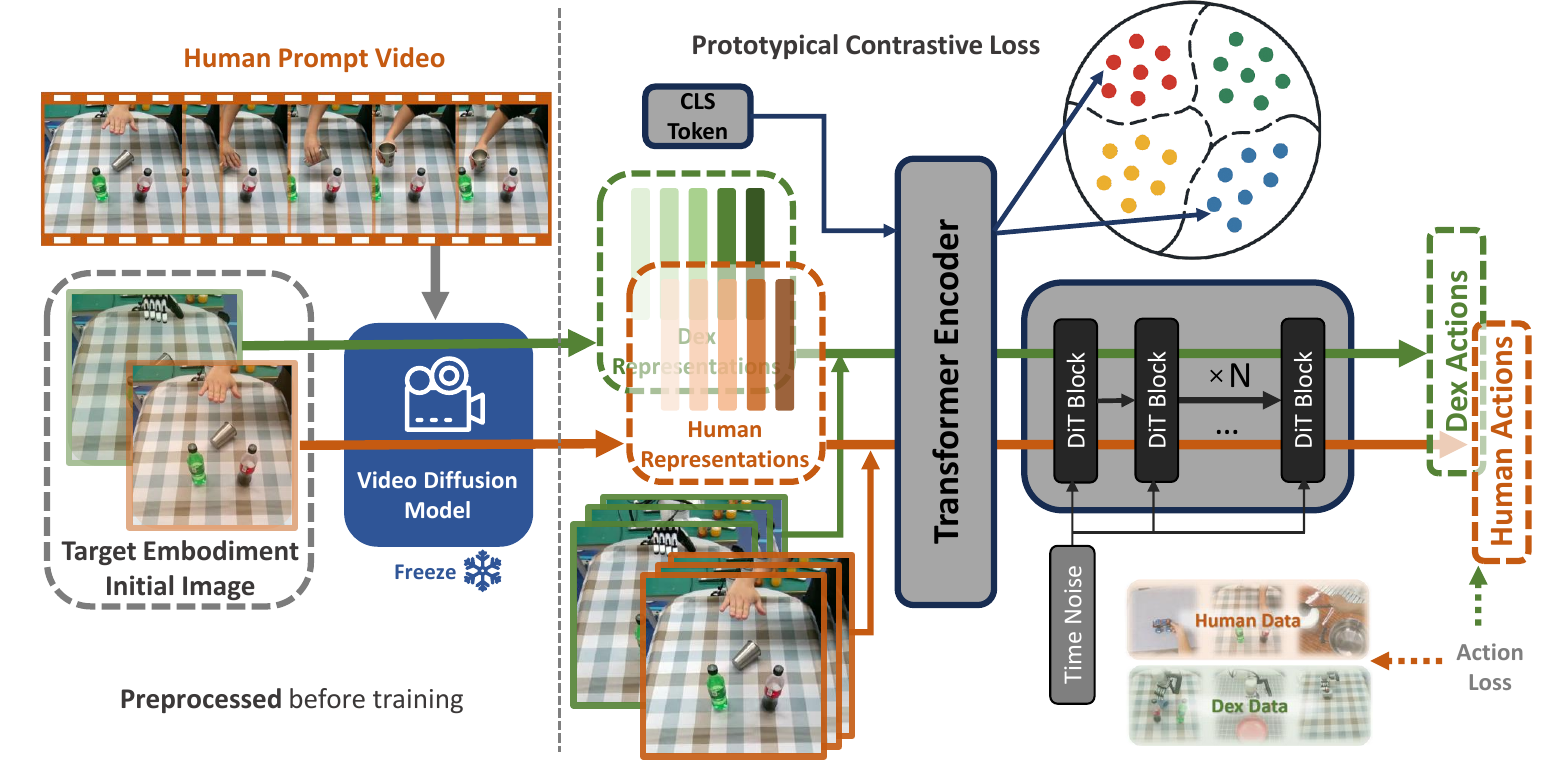}
    \caption{%
        \begin{minipage}{0.8\textwidth} 
        \textbf{Stage 2}: We use the diffusion model trained in the first stage to combine information 
        from human prompt videos with the target embodiment, generating an informative representation. 
        This representation, along with a shared action between human and robot and a prototypical 
        contrastive loss, enables the diffusion policy to learn task and skill information for both.
        \end{minipage}
    }
    \label{fig:stage2}
    \vspace*{-5mm}
\end{figure*}

\subsection{Skill Learning by Human Video-Action Pair Boosting}
\label{sec:human-action}

Training manipulation policies with dexterous hand data has become a key approach in robotic manipulation. However, collecting such data through teleoperation is time-consuming and costly. To address this, we reduce reliance on robotic hand data by leveraging human hand demonstrations to enhance manipulation capabilities. The availability of human hand data is virtually unlimited, with videos easily sourced from the internet or self-collected, requiring minimal time and infrastructure. We propose incorporating human hand demonstrations into a compatible format alongside teleoperated robotic hand data, which are then jointly trained using Imitation Learning. This forms the core methodology of our Stage-2 algorithm.

\textbf{Human Action Preprocessing: }The objective of human data processing is to establish correspondence between human demonstration data and robot motion data. Given that we only have third-person view RGB videos of human demonstrations, while the robot data contains both third-person view RGB videos and joint state information. The robot's end-effector state comprises two components: 6D Wrist pose (wrist position $\in$ $\mathbb{R}^3$ and wrist orientation $\in$ $\mathsf{SO}(3)$) and finger joint $\theta^{12}$. We employ the hand-tracking method WiLoR\cite{Potamias2024WiLoRE3} for hand localization in video frames and 3D hand mesh reconstruction, and the model outputs 21 right-hand keypoints $J_{h_i}\in\mathbb{R}^3$ for each frame. 

To extract the 6D wrist pose from 21 right-hand keypoints, we take the wrist keypoint $\mathbf{W}$ as the position. 
For orientation, a local right-handed coordinate frame is constructed using $\mathbf{W}$, the index-finger joint $\mathbf{I}$, and the middle-finger joint $\mathbf{M}$. 
The x-axis is defined along the direction from $\mathbf{W}$ to $\mathbf{M}$, the z-axis as the normalized normal of the plane formed by ($\mathbf{W}$, $\mathbf{I}$, $\mathbf{M}$) pointing toward the palm, and the y-axis is obtained via the right-hand rule. 
The resulting orthonormal basis forms the rotation matrix: $\mathbf{R} = [\mathbf{x}\ \mathbf{y}\ \mathbf{z}]$, which, together with $\mathbf{W}$, gives the complete 6D pose representation.

In addition, we compute the 3D position of the human hand in the camera coordinate system using the intrinsic parameters and transform it into the robot’s base coordinate system. To establish joint-level correspondence between the human and robot systems, motion retargeting is required, mapping the kinematic configuration of the human demonstrator’s hand to that of the robotic hand. Following AnyTeleop\cite{Qin2023AnyTeleopAG}, we formulate this problem as an optimization, thereby obtaining the data that map human hand joints in demonstration videos to the corresponding dexterous hand joints.

\textbf{Representation Conditioned Diffusion Policy: }Our framework employs a diffusion policy that models the conditional action distribution $p(a_t|s_t,z)$, where: $s_t = f_{resnet34}(o_t)$ $\in$ $\mathbb{R}^{1000}$ is the visual observation feature extracted by a pretrained ResNet-34, $z \in \mathbb{R}^{4096}$ denotes the stage-one representation. Here we define the \textbf{human representation} as the generated representation with a human video as the prompt and a human image as the initial frame, while the \textbf{robot representation} as the representation with a human video as the prompt and a robot image as the initial frame. $a_t$ $\in$ $\mathbb{R}^{19}$ is the action vector at time t. The diffusion process operates in the action space through:
\begin{equation}
    a_{t}^{i} = \sqrt{\alpha_i}a_0 + \sqrt{1 - \alpha_i}\epsilon, \epsilon \sim \mathcal{N}(0, \mathbf{I})
\end{equation}
while the reverse process learns to predict and remove the noise through a transformer-based denoising network $\epsilon_{\theta}$.

For dexterous hand control tasks, we decompose the action space into three distinct components: finger joint angles $a_{t}^{finger}\in\mathbb{R}^{12}$, wrist orientation $a_t^{rot}\in\mathsf{SO}(3)$, and wrist position $a_t^{pos}\in\mathbb{R}^3$. This decomposition enables specialized handling of each action modality through separate prediction heads in the denoising network. The training loss combines weighted component losses:
\begin{equation}
    \mathcal{L}_{action} = \lambda_f \mathcal{L}_{finger} + \lambda_r\mathcal{L}_{rotation}+\lambda_p\mathcal{L}_{postion}
\end{equation}
where:
\begin{equation}
    \mathcal{L}_* = \mathbb{E}_{i,a_0,\epsilon}[||\epsilon_*-\epsilon_{\theta}(a_t^{i}, i, h_0)||^2]
\end{equation}
$h_0=[s_t,z]W_e+b_e$ is the concatenated input token, $W_e$ is the embedding matrix, $b_e$ is the bias. $\lambda_f,\lambda_r,\lambda_p$ are weighting coefficients determined through validation.

\subsection{\textbf{P}roto\textbf{D}iffusion \textbf{C}ontrastive \textbf{P}olicy(PDCP)}
Siamese Prototypical Contrastive Learning (SPCL)\cite{li2020prototypical, mo2022siamese} first groups feature embeddings into prototypes via K-Means and then applies a Siamese‐style metric loss to pull together embeddings within each prototype while pushing apart those from different prototypes, alongside a prototypical cross-entropy loss that treats prototype assignments as soft labels to sharpen each sample’s affinity to its cluster. By leveraging learned cluster structure rather than individual instances, SPCL mitigates false negatives and yields more stable positive sets, thereby enhancing semantic discriminability in self-supervised learning.

To adapt SPCL for diffusion policy training, we add a learnable clustering token to the input of our Diffusion Transformer (DiT) encoder and use the encoder’s output for that token—denoted $h$—as a prototype‐aware latent. During training we jointly optimize three losses: 

(1) the NT-Xent contrastive loss $\mathcal{L}_{\mathrm{contra}}$, which treats tasks sharing the same skill as positives and tasks of different skills as negatives (analogous to data‐augmented positives in vision); (2) a prototypical cross-entropy loss $\mathcal{L}_{\mathrm{proto}}$, which encourages each the learned latent $\mathbf{h}$ to align with the cross-entropy distribution over its K-Means prototype label, promoting tighter intra-prototype clustering and clearer inter-prototype separation.
And (3) a Siamese-style metric loss $\mathcal{L}_{\mathrm{metric}}$ at the prototype level, which further enforces proximity among same-prototype samples in metric space while repelling different-prototype samples, thus reducing semantic confusion. Due to space constraints, we omit the detailed formulations of $\mathcal{L}_{contra}$, $\mathcal{L}_{proto}$, and $\mathcal{L}_{metric}$, which are similar in spirit to those in \cite{li2020prototypical, mo2022siamese}, but adapted to our setting with modifications in sample construction.

The total loss for PDCP consist of the action loss and a weighted sum of the mentioned loss with coefficient $w_p, w_p, w_m$:
\begin{equation}
    \mathcal{L} = \mathcal{L}_{action} + w_c\mathcal{L}_{\mathrm{contra}} + w_p\mathcal{L}_{\mathrm{proto}} + w_m\mathcal{L}_{\mathrm{metric}}
\end{equation}
These objectives encourage task-discriminative and modality-aligned representations, improving cross-modal generalization.


\section{EXPERIMENTS}

In the following section, we present a comprehensive set of experiments to validate our proposed method. We deploy the algorithm on the dexterous-hand Xhand, assigning it 13 fundamentally distinct skills, such as pick-and-place, pressing, pouring, flipping cups, push operations, ball flicking and so on. These skills are further decomposed into more than 100 concrete tasks. We evaluate generalization across positions, scenes, objects, and, most critically, across skills. Our experiments are designed to address the following questions:
(1) Can cross-prediction enhance the performance, effectiveness, and representational capacity of the learned representation? (2) Is it feasible to use different video prompts in combination with a video generation model for skill extraction, and what is the resulting impact? (3) Can the extracted representation and DPCP enhance the position/scene/background generalization performance of the diffusion policy? (4) Does incorporating human action data improve the object generalization ability of the diffusion policy? (5) Can the final learned policy demonstrate the ability to generalize to previously unseen robot skills as well as exhibit "in-context learning" capabilities?



In our real-robot experiments, we conducted a systematic evaluation of the proposed policy. 
For Stage-1 representation learning, we used over 250,000 videos of human, gripper, and dexterous-hand demonstrations from RH20T\cite{fang2023rh20t}, Something-Something\cite{goyal2017something}, HOI4D\cite{Liu2022HOI4DA4}, Bridge\cite{walke2023bridgedata} and our in-house dexterous manipulation dataset collected via Apple Vision Pro teleoperation.

For Stage-2 policy learning, we employed about 7,000 human and dexterous-hand videos, including paired demonstrations collected in-house and additional human videos from HOI4D\cite{Liu2022HOI4DA4}. 
The ratio of human to dexterous-hand videos was about 5:1, ensuring the dataset remained sparse in robot data but rich in human demonstrations.



At inference, our method extracts task representations from human prompt videos, fuses them with the robot’s real-time observations, and then feeds them into the stage-2 policy to generate actions. We use the DiT from RDT-1B\cite{Liu2024RDT1BAD} as our policy network to generate continuous actions in a unified action space conditioned on multimodal inputs and embodiment embeddings. The actions are executed by the robot’s built-in position controller at 10 Hz. We compare our approach against the following baselines:

\begin{itemize}
    \item \textbf{Representation+Robot.} Policies trained solely on robot data without incorporating any human demonstrations.
    \item \textbf{Language+Robot+Human.} Policies conditioned on CLIP-encoded task labels (e.g., “pour water”) as inputs, rather than on learned skill representations.
    \item \textbf{Language-conditioned Diffusion Policy.} The language input is processed with a CLIP text encoder and subsequently incorporated as a conditional input to the Diffusion Policy\cite{Chi2023DiffusionPV}, which then generates the corresponding action.
    \item \textbf{Representation+Robot+Human.} Policies jointly trained on both robot and human data, optimized using learned skill representations.
    \item \textbf{XSkill\cite{xu2023xskill}.} In the first-stage representation learning, XSkill employed the same human and robot datasets as ours. However, due to methodological constraints, its second-stage training relied solely on robot data.
    \item \textbf{UniSkill\cite{kim2025uniskill}.} In UniSkill\cite{kim2025uniskill}, the first stage trains Inverse and Forward Skill Dynamics using the same human and robot data as our method. The second stage learns the policy with robot data, and inference is conditioned on human videos as prompts.
\end{itemize}


For each algorithm and each generalization level, we conducted 60-80 trials to ensure statistical robustness. To guarantee a fair comparison, all methods were evaluated under identical experimental conditions: For each evaluation task,
the generalization level (position/scene/background\&object\&skill) and the prompt (either human video prompts or corresponding textual descriptions) are the same. Performance was measured by two metrics: Success Rate (SR), defined as the fraction of successful trials, and Task Score, which quantifies task progress by decomposing long-horizon tasks into subtasks, assigning partial credit, and normalizing the cumulative score to 100.

\begin{figure}[!t]
    \centering
    \includegraphics[width=1.00\linewidth]{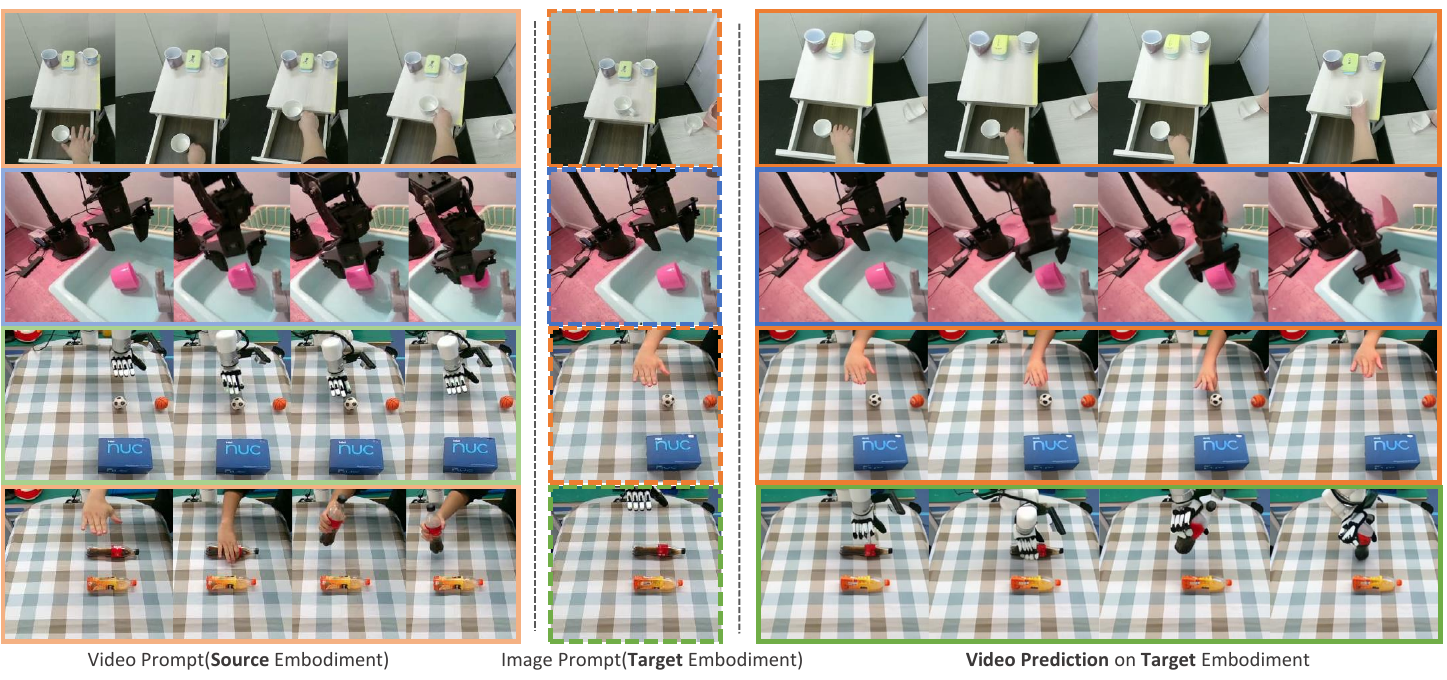}
    \vspace*{-7mm}
    \captionof{figure}{Examples of Cross-Prediction Video Generation.}
    \label{fig:video gen}
    \vspace*{-5mm}
\end{figure}

\subsection{Evaluation for Cross Prediction Video Generation}

Our cross-prediction mechanism extracts the task performed by the source embodiment from the prompt video and, using the scene context and target embodiment information form another given image, generates a video of the target embodiment executing that task.
As shown in Fig.\ref{fig:video gen}, we show two examples of same-embodiment prediction alongside with two examples of cross-embodiment prediction. The generated videos faithfully capture the task details from the prompt, and in the cross-embodiment cases, the generated target embodiment motion and scene context align precisely with the original task. This demonstrates that our model has learned not only the physical dynamics of the world but also learned how to transfer task and skill across different embodiments.


\begin{figure*}[b!] 
    \vspace*{-6mm}
    \centering
    \subfloat[t-SNE visualization of learned representations.]{%
        \includegraphics[width=0.49\textwidth]{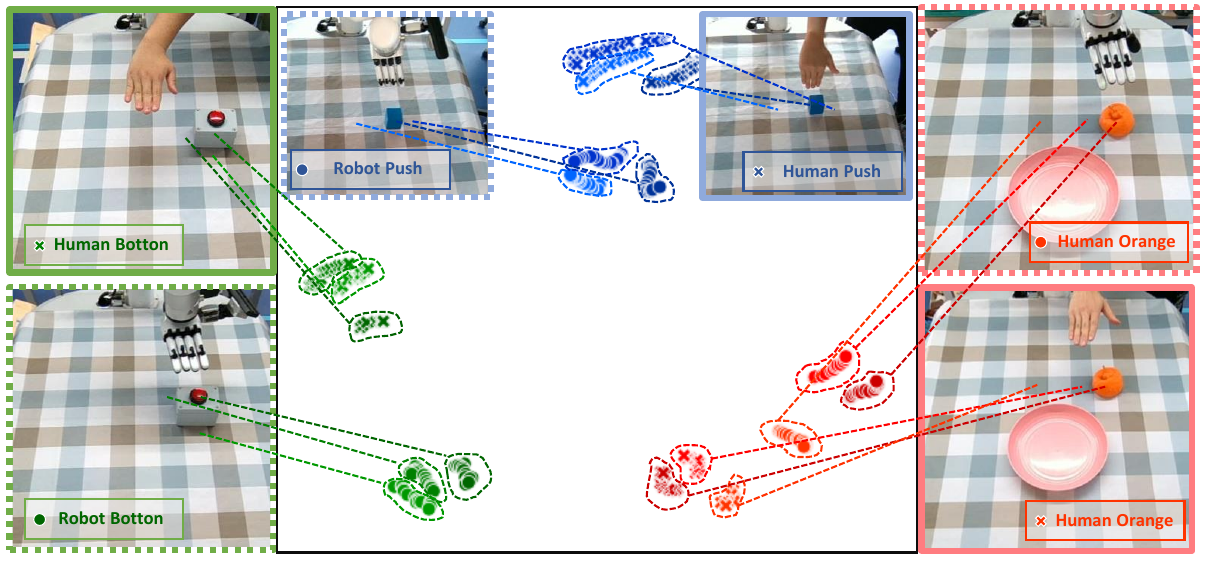}%
        \label{fig:t-SNE}
    }
    \hfill
    \subfloat[Vector properties of learned representations.]{%
        \includegraphics[width=0.49\textwidth]{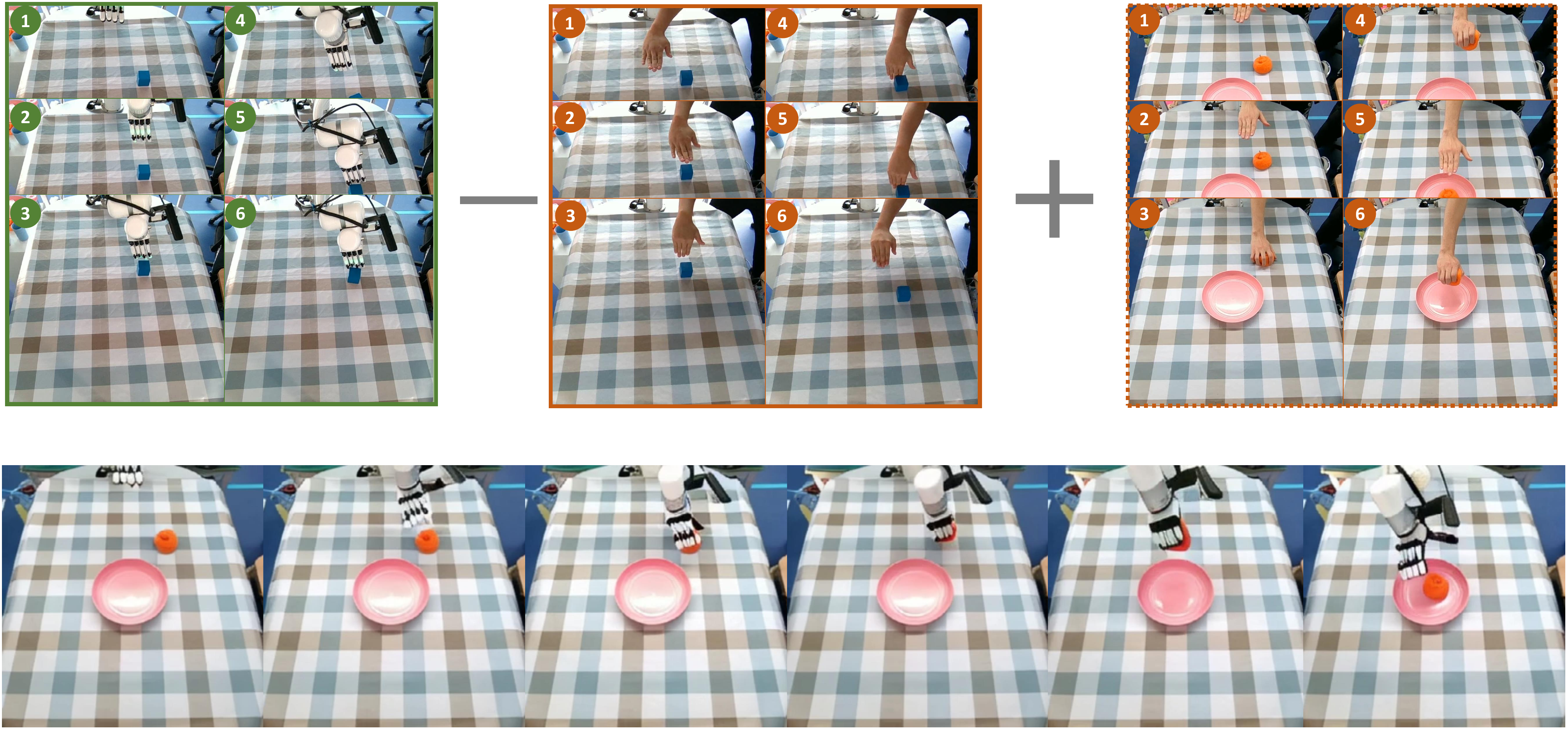}%
        \label{fig:latent validation}
    }
    \caption{Evaluation for the Learned Representations}
    \label{fig:eval rep'}
\end{figure*}

\subsection{Evaluation for the Learned Representations}
As shown in previous experiments, applying cross-prediction enables the video generation model to convert between human and robot embodiments while retaining its original capabilities. However, our goal is to extract the task information performed by the source embodiment from the prompt video using the cross-prediction fine-tuned SVD, and then transfer that information to the target embodiment. So, how do we evaluate and verify the performance of the representations extracted by the video generation model? We conducted the following experiments:


\subsubsection{\textbf{t-SNE Visualization of the Learned Representation}}

We conducted a t-SNE visualization experiment. We applied t-SNE to reduce the dimensionality of the skills extracted from the \textbf{human representation}(with a human video as the prompt and a human image as the initial frame) and the \textbf{robot representation} (with a human video as the prompt and a robot image as the initial frame), and then visualized them on a two-dimensional plane. The results, shown in Fig.\ref{fig:t-SNE}, reveal that tasks of the same category are grouped by similar color schemes, with identical hues indicating representations at different time points. Dashed lines highlight the positions of target objects in the two-dimensional space. The plot demonstrates that representations can effectively distinguish tasks, with the same task’s representations clustering closely based on embodiment, while still distinguishing target object positions. Additionally, representations from different time steps within the same video are tightly grouped yet distinguishable, reflecting the capture of continuous temporal information. These results show that the learned representations not only encode embodiment and task information but also capture object positioning and motion details, which is challenging to obtain with language prompts, showcasing the advantage of video prompts.


\subsubsection{\textbf{Vector Properties of the Learned Representations}} In word embeddings, a fascinating phenomenon is observed: these embeddings exhibit an intrinsic vector structure. For example, taking the representation of “king,” subtracting that of “queen,” and then adding the representation of “nurse” yields that of “doctor.” This arithmetic property is celebrated as both elegant and intriguing. Thus, we aim to explore whether our learned representations possess a similar vectorial nature. We subtracted the human representation from the corresponding robot representation for the "push blue box" task and then added the human representation for the "grasp orange" task. Using this composite representation, we generated a video that, as shown in the Fig.\ref{fig:latent validation}, ultimately depicted a robot executing the "grasp orange" task. Moreover, we conducted tests on other tasks—such as "push button" and "pour water" and observed similarly intriguing results. This demonstrates that our learned representations indeed exhibit notable vector properties.


\begin{table}
\vspace*{1mm}
\centering
\caption{Success Rate (SR) and Score metrics on position, scene, and background generalization}
\vspace*{-3mm}
\label{tab:results}
\begin{tabular}{l*{3}{cc}}
\toprule
\multirow{2}{*}{Method} & \multicolumn{2}{c}{Position} & \multicolumn{2}{c}{Scene} & \multicolumn{2}{c}{Background} \\
\cmidrule(lr){2-3}\cmidrule(lr){4-5}\cmidrule(lr){6-7}
 & SR & Score & SR & Score & SR & Score \\
\midrule
Xskill\cite{xu2023xskill} & 0.31 & 60.1 & 0.27 & 46.4 & 0.18 & 47.3\\
\addlinespace[3pt]
Language+R.+H. & 0.58 & 69.5 & 0.56 & 55.6 & 0.36 & 50.9 \\
\addlinespace[3pt]
Repr.+R. & 0.68 & 77.9 & 0.67 & 81.1 & 0.55 & 69.1 \\
\addlinespace[3pt]
Repr.+R.+H. & 0.74 & 81.6 & 0.67 & 84.4& 0.64 & 74.5 \\
\addlinespace[3pt]
\makecell{\textbf{Repr.+R.+H.+DPCP} \\ \textbf{(Ours)}} & \textbf{0.79} & \textbf{85.3} & \textbf{0.69} & \textbf{88.9} & \textbf{0.73} & \textbf{80.9}\\
\bottomrule
\end{tabular}
\vspace*{-5mm}
\end{table}

\subsection{Position, Scene, Background Generalization}

To assess the generalization capabilities of our policy, we conducted comprehensive testing across three critical dimensions. \textbf{Positional Generalization} means randomly relocating target objects within the workspace. \textbf{Scene Generalization} means randomly changing or replacing inoperable objects within the workspace. \textbf{Background Generalization} means testing under different background (tablecloth, ...).

As shown by the comparative results in Table~\ref{tab:results}, our method consistently outperforms the four baseline approaches across all three generalization scenarios. A further comparison between our Stage-1 representation-guided policy and representation learning methods conditioned on language or based on XSkill\cite{xu2023xskill} reveals the fundamental advantages of our learned representations in encoding richer task-relevant information. This advantage arises from the integration of end-effector positional information and partial operational knowledge within our representation learning framework. Our analysis also demonstrates that the proposed PDCP algorithm effectively mitigates mis-grasping issues when handling closely spaced objects. We argue that while the Stage-1 representation primarily captures spatial variations, the PDCP module provides complementary functionality by enabling task-aware feature clustering.


\begin{figure}[h]
    \centering
    \vspace*{-3mm}
    \includegraphics[width=0.47\textwidth]{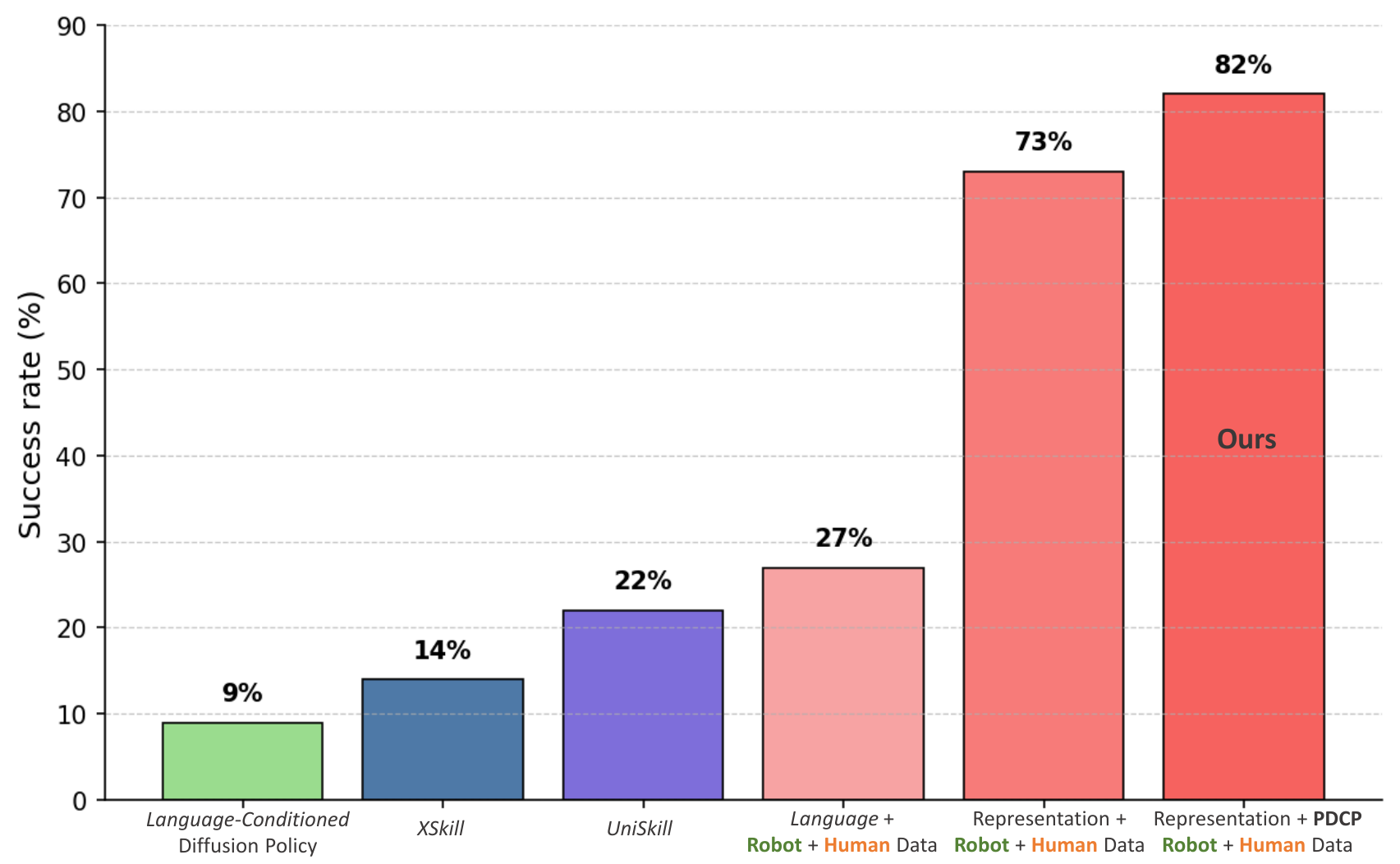}%
    \vspace*{-3mm}
    \captionof{figure}{Evaluation on Obj Generalization}
    \vspace{-6mm}
    \label{fig:algorithm_performance}
\end{figure}

\subsection{Object Generalization}
To investigate the complementary benefits of our representation learning framework and human demonstration data, we design object-level generalization experiments to evaluate their individual and combined effects. While our representations perform well across all generalization scenarios, these experiments isolate and quantify the contribution of human data with novel object instances.

We define a \textbf{New Object} as one that does not appear in robot data during the second-stage training but is present in human demonstrations. Accordingly, for our method and the baselines that incorporate human videos in stage-2 training (Representation+Robot+Human Data and Language+Robot+Human Data), we use all available robot data excluding demonstrations involving the new object, together with the full set of human data, which includes demonstrations of that object. In contrast, for the baselines that train solely on robot videos in stage-2 (Language-Conditioned Diffusion Policy, XSkill\cite{xu2023xskill}, and UniSkill\cite{kim2025uniskill}), we use only the robot data while excluding demonstrations involving that new object.

As shown in Figure \ref{fig:algorithm_performance}, we evaluate four distinct tasks involving unseen objects and compare our approach against baseline methods. At the new object generalization level,
compared to training without human demonstrations, the success rate on new objects improves significantly, underscoring the role of human data in transferring knowledge of object affordances and manipulation strategies. 
Our representation further complements this by supplying spatial cues and partial operational knowledge for tasks absent from the robot training set, while human demonstrations offer complementary insights into novel objects and strategies.
In contrast, the XSkill and UniSkill baselines cannot leverage human actions during the second-stage policy learning process, leaving them unable to transfer knowledge from human demonstrations and resulting in poorer performance on unseen-object tasks.

\begin{figure}[h]
    \centering
    \vspace*{-3mm}
    \includegraphics[width=0.49\textwidth]{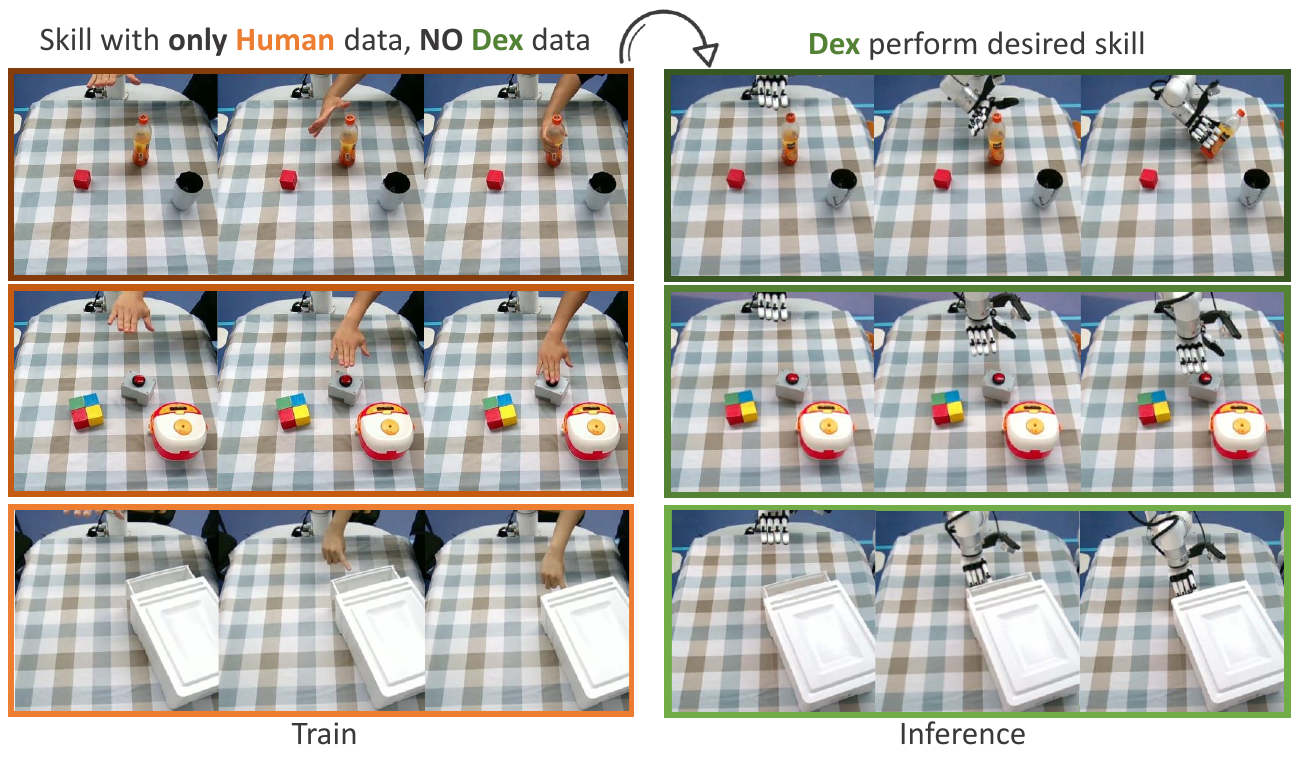}%
    \vspace*{-5mm}
    \captionof{figure}{Unseen Skill transfers}
    \label{fig:new_skill}
    \vspace*{-5mm}
\end{figure}

\subsection{Skill Generalization}

Having shown that our policy can generalize to objects absent from the robot training dataset, we next examine whether it can also acquire entirely new skills missing from the original robot data. To this end, we evaluate four novel tasks: grasping an orange juice bottle, pressing a button, closing a drawer, and opening a carton. For each skill, the policy was trained on all available robot trajectories \textbf{excluding those involving the target skill}, while jointly incorporating the full set of human demonstrations that included the skill. 
For instance, in the case of pressing skill, all robot trajectories involving pressing action were excluded, whereas the human dataset retained demonstrations for all pressing tasks.
To improve visual alignment between human and robot demonstrations, we further applied a pre-processing trick: in human videos, a solid circle was overlaid on the detected wrist position with its size inversely proportional to camera distance; in robot videos, the wrist was annotated in the same way. As illustrated in Figure~\ref{fig:new_skill}, our approach enables the policy to “watch” human demonstrations (left) and successfully execute previously unseen skills on the robot (right), showcasing a degree of video-based "in-context learning".

For baselines that incorporate human videos in stage-2 training (Language+R.+H. and Repr.+R.+H.+DPCP (w/o Circle)), we use all available robot data excluding the target skill, along with the full set of human demonstrations. For baselines relying solely on robot videos (XSkill\cite{xu2023xskill}, UniSkill\cite{kim2025uniskill}, and Language-Conditioned Diffusion Policy), only robot data excluding the target skill is used.


As shown in Table~\ref{tab:new_skill}, our method consistently outperforms these baselines. Unlike XSkill and UniSkill, which cannot exploit human data in stage-2, our approach effectively leverages human demonstrations to acquire novel skills and improve execution on unseen skills. Moreover, adding circle tags reduces the human–robot visual gap and provides position-aware cues, yielding more accurate actions. These results underscore the importance of human demonstrations and image-level alignment, and validate the effectiveness of our method in enabling robots to generalize to entirely unseen skills.

\begin{table}[t]
\vspace{1mm}
\centering
\caption{Performance on novel skill generation tasks in terms of Success Rate (SR) and Score.}

\label{tab:new_skill}
\scriptsize
\setlength{\tabcolsep}{2pt}        
\renewcommand{\arraystretch}{1.2} 
\resizebox{\linewidth}{!}{
\begin{tabular}{l*{8}{>{\centering\arraybackslash}p{0.95cm}}}
\toprule
\multirow{2}{*}{Method} 
& \multicolumn{2}{c}{Grasp Bottle} 
& \multicolumn{2}{c}{Press Button} 
& \multicolumn{2}{c}{Close Drawer} 
& \multicolumn{2}{c}{Open Carton} \\
\cmidrule(lr){2-3}\cmidrule(lr){4-5}\cmidrule(lr){6-7}\cmidrule(lr){8-9}
& SR & Score & SR & Score & SR & Score & SR & Score \\
\midrule
XSkill~\cite{xu2023xskill}                          & 0.00 & 0.0 & 0.00 & 0.0 & 0.00 & 0.0 & 0.00 & 0.0 \\
UniSkill~\cite{kim2025uniskill}                          & 0.00 & 0.0 & 0.00 & 0.0 & 0.00 & 0.0 & 0.00 & 0.0 \\
\makecell{Language-conditioned Diffusion Policy}                       & 0.00 & 0.0 & 0.00 & 0.0 & 0.00 & 0.0 & 0.00 & 0.0 \\
Language+R.+H.                                      & 0.18 & 27.3 & 0.22 & 28.9 & 0.25 & 41.3 & 0.08 & 17.7 \\
Repr.+R.+H.+DPCP(w/o Circle)         & 0.27 & 43.3 & 0.25 & 36.3 & 0.35 & 49.4 & 0.23 & 36.9 \\
\textbf{Repr.+R.+H.+DPCP}\textbf{(Ours)} & \textbf{0.47} & \textbf{65.3} & \textbf{0.56} & \textbf{63.8} & \textbf{0.65} & \textbf{66.5} & \textbf{0.46} & \textbf{50.8} \\
\bottomrule
\end{tabular}
}
\vspace{-5mm}
\end{table}

\section{CONCLUSIONS}
This work fully exploits human data across both the representation learning and diffusion-based policy stages. By using human videos as prompts, which provide substantially richer information than language inputs, our approach achieves superior performance on dexterous hand manipulation tasks.



\bibliographystyle{IEEEtran}
\bibliography{cite}

@IEEEtranBSTCTL{IEEEexample:BSTcontrol,
  CTLuse_forced_etal       = "yes",
  CTLmax_names_forced_etal = "3",
  CTLnames_show_etal       = "1",
}

@inproceedings{xiong2021learning,
  title={Learning by watching: Physical imitation of manipulation skills from human videos},
  author={Xiong, Haoyu and Li, Quanzhou and Chen, Yun-Chun and Bharadhwaj, Homanga and Sinha, Samarth and Garg, Animesh},
  booktitle={2021 IEEE/RSJ International Conference on Intelligent Robots and Systems (IROS)},
  pages={7827--7834},
  year={2021},
  organization={IEEE}
}

@inproceedings{chane2023learning,
  title={Learning video-conditioned policies for unseen manipulation tasks},
  author={Chane-Sane, Elliot and Schmid, Cordelia and Laptev, Ivan},
  booktitle={2023 IEEE International Conference on Robotics and Automation (ICRA)},
  pages={909--916},
  year={2023},
  organization={IEEE}
}

@article{jain2024vid2robot,
  title={Vid2robot: End-to-end video-conditioned policy learning with cross-attention transformers},
  author={Jain, Vidhi and Attarian, Maria and Joshi, Nikhil J and Wahid, Ayzaan and Driess, Danny and Vuong, Quan and Sanketi, Pannag R and Sermanet, Pierre and Welker, Stefan and Chan, Christine and others},
  journal={arXiv preprint arXiv:2403.12943},
  year={2024}
}

@article{qian2024contrast,
  title={Contrast, Imitate, Adapt: Learning Robotic Skills From Raw Human Videos},
  author={Qian, Zhifeng and You, Mingyu and Zhou, Hongjun and Xu, Xuanhui and Fu, Hao and Xue, Jinzhe and He, Bin},
  journal={IEEE Transactions on Automation Science and Engineering},
  year={2024},
  publisher={IEEE}
}

@inproceedings{xu2023xskill,
  title={Xskill: Cross embodiment skill discovery},
  author={Xu, Mengda and Xu, Zhenjia and Chi, Cheng and Veloso, Manuela and Song, Shuran},
  booktitle={Conference on robot learning},
  pages={3536--3555},
  year={2023},
  organization={PMLR}
}

@inproceedings{chang2023one,
  title={One-shot visual imitation via attributed waypoints and demonstration augmentation},
  author={Chang, Matthew and Gupta, Saurabh},
  booktitle={2023 IEEE International Conference on Robotics and Automation (ICRA)},
  pages={5055--5062},
  year={2023},
  organization={IEEE}
}

@inproceedings{xiong2023robotube,
  title={Robotube: Learning household manipulation from human videos with simulated twin environments},
  author={Xiong, Haoyu and Fu, Haoyuan and Zhang, Jieyi and Bao, Chen and Zhang, Qiang and Huang, Yongxi and Xu, Wenqiang and Garg, Animesh and Lu, Cewu},
  booktitle={Conference on Robot Learning},
  pages={1--10},
  year={2023},
  organization={PMLR}
}

@article{chen2024vividex,
  title={Vividex: Learning vision-based dexterous manipulation from human videos},
  author={Chen, Zerui and Chen, Shizhe and Arlaud, Etienne and Laptev, Ivan and Schmid, Cordelia},
  journal={arXiv preprint arXiv:2404.15709},
  year={2024}
}

@inproceedings{qin2022dexmv,
  title={Dexmv: Imitation learning for dexterous manipulation from human videos},
  author={Qin, Yuzhe and Wu, Yueh-Hua and Liu, Shaowei and Jiang, Hanwen and Yang, Ruihan and Fu, Yang and Wang, Xiaolong},
  booktitle={European Conference on Computer Vision},
  pages={570--587},
  year={2022},
  organization={Springer}
}

@article{wang2024dexcap,
  title={Dexcap: Scalable and portable mocap data collection system for dexterous manipulation},
  author={Wang, Chen and Shi, Haochen and Wang, Weizhuo and Zhang, Ruohan and Fei-Fei, Li and Liu, C Karen},
  journal={arXiv preprint arXiv:2403.07788},
  year={2024}
}

@article{kannan2023deft,
  title={Deft: Dexterous fine-tuning for real-world hand policies},
  author={Kannan, Aditya and Shaw, Kenneth and Bahl, Shikhar and Mannam, Pragna and Pathak, Deepak},
  journal={arXiv preprint arXiv:2310.19797},
  year={2023}
}

@article{yuan2024general,
  title={General flow as foundation affordance for scalable robot learning},
  author={Yuan, Chengbo and Wen, Chuan and Zhang, Tong and Gao, Yang},
  journal={arXiv preprint arXiv:2401.11439},
  year={2024}
}

@inproceedings{ju2024robo,
  title={Robo-abc: Affordance generalization beyond categories via semantic correspondence for robot manipulation},
  author={Ju, Yuanchen and Hu, Kaizhe and Zhang, Guowei and Zhang, Gu and Jiang, Mingrun and Xu, Huazhe},
  booktitle={European Conference on Computer Vision},
  pages={222--239},
  year={2024},
  organization={Springer}
}

@article{wang2023mimicplay,
  title={Mimicplay: Long-horizon imitation learning by watching human play},
  author={Wang, Chen and Fan, Linxi and Sun, Jiankai and Zhang, Ruohan and Fei-Fei, Li and Xu, Danfei and Zhu, Yuke and Anandkumar, Anima},
  journal={arXiv preprint arXiv:2302.12422},
  year={2023}
}

@article{smith2019avid,
  title={Avid: Learning multi-stage tasks via pixel-level translation of human videos},
  author={Smith, Laura and Dhawan, Nikita and Zhang, Marvin and Abbeel, Pieter and Levine, Sergey},
  journal={arXiv preprint arXiv:1912.04443},
  year={2019}
}

@article{wen2023any,
  title={Any-point trajectory modeling for policy learning},
  author={Wen, Chuan and Lin, Xingyu and So, John and Chen, Kai and Dou, Qi and Gao, Yang and Abbeel, Pieter},
  journal={arXiv preprint arXiv:2401.00025},
  year={2023}
}

@article{gu2023rt,
  title={Rt-trajectory: Robotic task generalization via hindsight trajectory sketches},
  author={Gu, Jiayuan and Kirmani, Sean and Wohlhart, Paul and Lu, Yao and Arenas, Montserrat Gonzalez and Rao, Kanishka and Yu, Wenhao and Fu, Chuyuan and Gopalakrishnan, Keerthana and Xu, Zhuo and others},
  journal={arXiv preprint arXiv:2311.01977},
  year={2023}
}

@article{kareer2024egomimic,
  title={Egomimic: Scaling imitation learning via egocentric video},
  author={Kareer, Simar and Patel, Dhruv and Punamiya, Ryan and Mathur, Pranay and Cheng, Shuo and Wang, Chen and Hoffman, Judy and Xu, Danfei},
  journal={arXiv preprint arXiv:2410.24221},
  year={2024}
}

@article{ye2024latent,
  title={Latent action pretraining from videos},
  author={Ye, Seonghyeon and Jang, Joel and Jeon, Byeongguk and Joo, Sejune and Yang, Jianwei and Peng, Baolin and Mandlekar, Ajay and Tan, Reuben and Chao, Yu-Wei and Lin, Bill Yuchen and others},
  journal={arXiv preprint arXiv:2410.11758},
  year={2024}
}

@article{liang2024dreamitate,
  title={Dreamitate: Real-world visuomotor policy learning via video generation},
  author={Liang, Junbang and Liu, Ruoshi and Ozguroglu, Ege and Sudhakar, Sruthi and Dave, Achal and Tokmakov, Pavel and Song, Shuran and Vondrick, Carl},
  journal={arXiv preprint arXiv:2406.16862},
  year={2024}
}

@article{du2023learning,
  title={Learning universal policies via text-guided video generation},
  author={Du, Yilun and Yang, Sherry and Dai, Bo and Dai, Hanjun and Nachum, Ofir and Tenenbaum, Josh and Schuurmans, Dale and Abbeel, Pieter},
  journal={Advances in neural information processing systems},
  volume={36},
  pages={9156--9172},
  year={2023}
}

@article{bharadhwaj2024gen2act,
  title={Gen2act: Human video generation in novel scenarios enables generalizable robot manipulation},
  author={Bharadhwaj, Homanga and Dwibedi, Debidatta and Gupta, Abhinav and Tulsiani, Shubham and Doersch, Carl and Xiao, Ted and Shah, Dhruv and Xia, Fei and Sadigh, Dorsa and Kirmani, Sean},
  journal={arXiv preprint arXiv:2409.16283},
  year={2024}
}

@article{shridhar2024generative,
  title={Generative image as action models},
  author={Shridhar, Mohit and Lo, Yat Long and James, Stephen},
  journal={arXiv preprint arXiv:2407.07875},
  year={2024}
}

@article{Ho2020DenoisingDP,
  title={Denoising Diffusion Probabilistic Models},
  author={Jonathan Ho and Ajay Jain and P. Abbeel},
  journal={ArXiv},
  year={2020},
  volume={abs/2006.11239},
}

@article{Song2020ScoreBasedGM,
  title={Score-Based Generative Modeling through Stochastic Differential Equations},
  author={Yang Song and Jascha Narain Sohl-Dickstein and Diederik P. Kingma and Abhishek Kumar and Stefano Ermon and Ben Poole},
  journal={ArXiv},
  year={2020},
  volume={abs/2011.13456},
}

@article{Chi2023DiffusionPV,
  title={Diffusion Policy: Visuomotor Policy Learning via Action Diffusion},
  author={Cheng Chi and Siyuan Feng and Yilun Du and Zhenjia Xu and Eric Cousineau and Benjamin Burchfiel and Shuran Song},
  journal={ArXiv},
  year={2023},
  volume={abs/2303.04137},
}

@article{Pearce2023ImitatingHB,
  title={Imitating Human Behaviour with Diffusion Models},
  author={Tim Pearce and Tabish Rashid and Anssi Kanervisto and David Bignell and Mingfei Sun and Raluca Georgescu and Sergio Valcarcel Macua and Shan Zheng Tan and Ida Momennejad and Katja Hofmann and Sam Devlin},
  journal={ArXiv},
  year={2023},
  volume={abs/2301.10677},
}

@article{Team2024OctoAO,
  title={Octo: An Open-Source Generalist Robot Policy},
  author={Octo Model Team and Dibya Ghosh and Homer Rich Walke and Karl Pertsch and Kevin Black and Oier Mees and Sudeep Dasari and Joey Hejna and Tobias Kreiman and Charles Xu and Jianlan Luo and You Liang Tan and Pannag R. Sanketi and Quan Vuong and Ted Xiao and Dorsa Sadigh and Chelsea Finn and Sergey Levine},
  journal={ArXiv},
  year={2024},
  volume={abs/2405.12213},
}

@article{Reuss2024MultimodalDT,
  title={Multimodal Diffusion Transformer: Learning Versatile Behavior from Multimodal Goals},
  author={Moritz Reuss and {\"O}mer Erdinç Yagmurlu and Fabian Wenzel and Rudolf Lioutikov},
  journal={ArXiv},
  year={2024},
  volume={abs/2407.05996},
}

@article{Liu2024RDT1BAD,
  title={RDT-1B: a Diffusion Foundation Model for Bimanual Manipulation},
  author={Songming Liu and Lingxuan Wu and Bangguo Li and Hengkai Tan and Huayu Chen and Zhengyi Wang and Ke Xu and Hang Su and Jun Zhu},
  journal={ArXiv},
  year={2024},
  volume={abs/2410.07864},
  url={https://api.semanticscholar.org/CorpusID:273233386}
}

@article{Potamias2024WiLoRE3,
  title={WiLoR: End-to-end 3D Hand Localization and Reconstruction in-the-wild},
  author={Rolandos Alexandros Potamias and Jinglei Zhang and Jiankang Deng and Stefanos Zafeiriou},
  journal={ArXiv},
  year={2024},
  volume={abs/2409.12259},
  url={https://api.semanticscholar.org/CorpusID:272753318}
}

@article{Qin2023AnyTeleopAG,
  title={AnyTeleop: A General Vision-Based Dexterous Robot Arm-Hand Teleoperation System},
  author={Yuzhe Qin and Wei Yang and Binghao Huang and Karl Van Wyk and Hao Su and Xiaolong Wang and Yu-Wei Chao and Dieter Fox},
  journal={ArXiv},
  year={2023},
  volume={abs/2307.04577},
  url={https://api.semanticscholar.org/CorpusID:259367735}
}

@article{Reuss2023GoalConditionedIL,
  title={Goal-Conditioned Imitation Learning using Score-based Diffusion Policies},
  author={Moritz Reuss and Maximilian Xiling Li and Xiaogang Jia and Rudolf Lioutikov},
  journal={ArXiv},
  year={2023},
  volume={abs/2304.02532},
  url={https://api.semanticscholar.org/CorpusID:257952177}
}

@article{Brohan2022RT1RT,
  title={RT-1: Robotics Transformer for Real-World Control at Scale},
  author={Anthony Brohan and Noah Brown and Justice Carbajal and Yevgen Chebotar and Joseph Dabis and Chelsea Finn and Keerthana Gopalakrishnan and Karol Hausman and Alexander Herzog and Jasmine Hsu and Julian Ibarz and Brian Ichter and Alex Irpan and Tomas Jackson and Sally Jesmonth and Nikhil J. Joshi and Ryan C. Julian and Dmitry Kalashnikov and Yuheng Kuang and Isabel Leal and Kuang-Huei Lee and Sergey Levine and Yao Lu and Utsav Malla and Deeksha Manjunath and Igor Mordatch and Ofir Nachum and Carolina Parada and Jodilyn Peralta and Emily Perez and Karl Pertsch and Jornell Quiambao and Kanishka Rao and Michael S. Ryoo and Grecia Salazar and Pannag R. Sanketi and Kevin Sayed and Jaspiar Singh and Sumedh Anand Sontakke and Austin Stone and Clayton Tan and Huong Tran and Vincent Vanhoucke and Steve Vega and Quan Ho Vuong and F. Xia and Ted Xiao and Peng Xu and Sichun Xu and Tianhe Yu and Brianna Zitkovich},
  journal={ArXiv},
  year={2022},
  volume={abs/2212.06817},
  url={https://api.semanticscholar.org/CorpusID:254591260}
}

@article{Brohan2023RT2VM,
  title={RT-2: Vision-Language-Action Models Transfer Web Knowledge to Robotic Control},
  author={Anthony Brohan{\relax} and Noah Brown and Justice Carbajal and Yevgen Chebotar and Krzysztof Choromanski and Tianli Ding and Danny Driess and Kumar Avinava Dubey and Chelsea Finn and Peter R. Florence and Chuyuan Fu and Montse Gonzalez Arenas and Keerthana Gopalakrishnan and Kehang Han and Karol Hausman and Alexander Herzog and Jasmine Hsu and Brian Ichter and Alex Irpan and Nikhil J. Joshi and Ryan C. Julian and Dmitry Kalashnikov and Yuheng Kuang and Isabel Leal and Sergey Levine and Henryk Michalewski and Igor Mordatch and Karl Pertsch and Kanishka Rao and Krista Reymann and Michael S. Ryoo and Grecia Salazar and Pannag R. Sanketi and Pierre Sermanet and Jaspiar Singh and Anikait Singh and Radu Soricut and Huong Tran and Vincent Vanhoucke and Quan Ho Vuong and Ayzaan Wahid and Stefan Welker and Paul Wohlhart and Ted Xiao and Tianhe Yu and Brianna Zitkovich},
  journal={ArXiv},
  year={2023},
  volume={abs/2307.15818},
  url={https://api.semanticscholar.org/CorpusID:260293142}
}

@article{mo2022siamese,
  title={Siamese prototypical contrastive learning},
  author={Mo, Shentong and Sun, Zhun and Li, Chao},
  journal={arXiv preprint arXiv:2208.08819},
  year={2022}
}

@article{li2020prototypical,
  title={Prototypical contrastive learning of unsupervised representations},
  author={Li, Junnan and Zhou, Pan and Xiong, Caiming and Hoi, Steven CH},
  journal={arXiv preprint arXiv:2005.04966},
  year={2020}
}

@article{li2024learning,
  title={Learning multimodal behaviors from scratch with diffusion policy gradient},
  author={Li, Steven and Krohn, Rickmer and Chen, Tao and Ajay, Anurag and Agrawal, Pulkit and Chalvatzaki, Georgia},
  journal={Advances in Neural Information Processing Systems},
  volume={37},
  pages={38456--38479},
  year={2024}
}

@article{wang2024diffusion,
  title={Diffusion actor-critic with entropy regulator},
  author={Wang, Yinuo and Wang, Likun and Jiang, Yuxuan and Zou, Wenjun and Liu, Tong and Song, Xujie and Wang, Wenxuan and Xiao, Liming and Wu, Jiang and Duan, Jingliang and others},
  journal={Advances in Neural Information Processing Systems},
  volume={37},
  pages={54183--54204},
  year={2024}
}

@article{jain2024sampling,
  title={Sampling from Energy-based Policies using Diffusion},
  author={Jain, Vineet and Akhound-Sadegh, Tara and Ravanbakhsh, Siamak},
  journal={arXiv preprint arXiv:2410.01312},
  year={2024}
}

@article{zhang2024entropy,
  title={Entropy-regularized diffusion policy with q-ensembles for offline reinforcement learning},
  author={Zhang, Ruoqi and Luo, Ziwei and Sj{\"o}lund, Jens and Sch{\"o}n, Thomas and Mattsson, Per},
  journal={Advances in Neural Information Processing Systems},
  volume={37},
  pages={98871--98897},
  year={2024}
}

@article{Xie2025Human2RobotLR,
  title={Human2Robot: Learning Robot Actions from Paired Human-Robot Videos},
  author={Sicheng Xie and Haidong Cao and Zejia Weng and Zhen Xing and Shiwei Shen and Jiaqi Leng and Xipeng Qiu and Yanwei Fu and Zuxuan Wu and Yu-Gang Jiang},
  journal={ArXiv},
  year={2025},
  volume={abs/2502.16587},
  url={https://api.semanticscholar.org/CorpusID:276575296}
}

@article{kim2025uniskill,
  title={UniSkill: Imitating Human Videos via Cross-Embodiment Skill Representations},
  author={Kim, Hanjung and Kang, Jaehyun and Kang, Hyolim and Cho, Meedeum and Kim, Seon Joo and Lee, Youngwoon},
  journal={arXiv preprint arXiv:2505.08787},
  year={2025}
}

@article{Liu2022HOI4DA4,
  title={HOI4D: A 4D Egocentric Dataset for Category-Level Human-Object Interaction},
  author={Yunze Liu and Yun Liu and Chen Jiang and Zhoujie Fu and Kangbo Lyu and Weikang Wan and Hao Shen and Bo-Hua Liang and He Wang and Li Yi},
  journal={2022 IEEE/CVF Conference on Computer Vision and Pattern Recognition (CVPR)},
  year={2022},
  pages={20981-20990},
  url={https://api.semanticscholar.org/CorpusID:247222786}
}

@article{fang2023rh20t,
  title={Rh20t: A comprehensive robotic dataset for learning diverse skills in one-shot},
  author={Fang, Hao-Shu and Fang, Hongjie and Tang, Zhenyu and Liu, Jirong and Wang, Chenxi and Wang, Junbo and Zhu, Haoyi and Lu, Cewu},
  journal={arXiv preprint arXiv:2307.00595},
  year={2023}
}

@inproceedings{goyal2017something,
  title={The" something something" video database for learning and evaluating visual common sense},
  author={Goyal, Raghav and Ebrahimi Kahou, Samira and Michalski, Vincent and Materzynska, Joanna and Westphal, Susanne and Kim, Heuna and Haenel, Valentin and Fruend, Ingo and Yianilos, Peter and Mueller-Freitag, Moritz and others},
  booktitle={Proceedings of the IEEE international conference on computer vision},
  pages={5842--5850},
  year={2017}
}

@inproceedings{walke2023bridgedata,
  title={Bridgedata v2: A dataset for robot learning at scale},
  author={Walke, Homer Rich and Black, Kevin and Zhao, Tony Z and Vuong, Quan and Zheng, Chongyi and Hansen-Estruch, Philippe and He, Andre Wang and Myers, Vivek and Kim, Moo Jin and Du, Max and others},
  booktitle={Conference on Robot Learning},
  pages={1723--1736},
  year={2023},
  organization={PMLR}
}
\clearpage



\end{document}